\documentclass[sigconf]{acmart}

\AtBeginDocument{%
  \providecommand\BibTeX{{%
    \normalfont B\kern-0.5em{\scshape i\kern-0.25em b}\kern-0.8em\TeX}}}

\setcopyright{acmcopyright}
\copyrightyear{2022}
\acmYear{2022}
\acmDOI{XXXXXXX.XXXXXXX}

\acmConference[SIGGRAPH '22 Posters]{In Special Interest Group on Computer Graphics and Interactive
Techniques Conference Posters}{August 8--11,2022}{Vancouver, CA}
\acmPrice{15.00}
\acmISBN{978-1-4503-XXXX-X/18/06}

\begin{document}

\title{Evaluating the Quality of a Synthesized Motion with the Fréchet Motion Distance}

\author{Antoine Maiorca}
\email{antoine.maiorca@umons.ac.be}
\orcid{0000-0001-5729-700X}
\affiliation{%
  \institution{ISIA Lab, University of Mons}
  \streetaddress{31 Boulevard Dolez}
  \city{Mons}
  \country{Belgium}
  \postcode{7000}
}

\author{Youngwoo Yoon}
\email{youngwoo@etri.re.kr}
\affiliation{%
  \institution{Electronics and Telecommunications Research Institute}
  \streetaddress{218, Gajeong-ro, Yuseong-gu}
  \city{Daejeon}
  \country{South Korea}}

\author{Thierry Dutoit}
\email{thierry.dutoit@umons.ac.be}
\affiliation{%
  \institution{ISIA Lab, University of Mons}
  \streetaddress{31 Boulevard Dolez}
  \city{Mons}
  \country{Belgium}
}

\keywords{Objective Motion Evaluation, Animation, Neural Networks}

\begin{teaserfigure}
  \includegraphics[width=\textwidth, height=4cm]{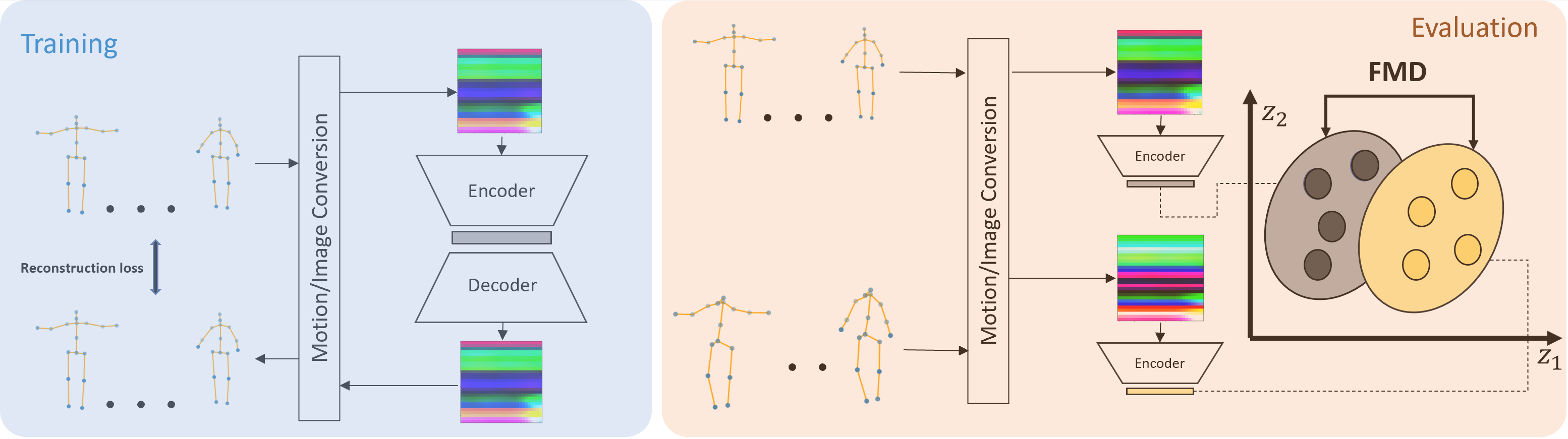}
  \caption{Overview of the FMD metric. During training process, the autoencoder learns a compressed motion feature space $z$ by reconstructing the input motion. Then, the encoder computes the latent space from the ground truth and synthetic motion dataset. Finally, the Fréchet distance is measured between the two latent spaces to evaluate the quality and diversity of the motion samples from the synthetic dataset.}
  \Description{}
  \label{fig:overview}
\end{teaserfigure}

\maketitle

\section{Introduction}
Motion synthesis is an active research topic in the Deep Learning community. It has various fields of application including character animation, humanoid robots and embodied conversational agents. Designing such algorithm is not a straightforward task and must ensure that the output motion are natural. It is however necessary to assess the quality of a synthesized animation to be able to compare algorithms' performance. Although that evaluations relying on subjective survey give satisfying estimation on the quality of the animation, gathering such group of people with defined requirements (level of expertise of subjects, e.g.) is expensive, time consuming, and having low reproducibility, which hinders fast development iterations. In this sense, a quantitative evaluation with an objective metric leverages those issues since it does not involve human in the loop. However, the design of such evaluation metric must ensure a strong correlation between it and the human perception on the motion quality. In the computer vision field, the metric called \textit{Fréchet Inception Distance} (FID) \cite{heusel2017gans} exhibits promising results in assessing synthesized images since it is sensitive to various image artifacts and penalizes the lack of diversity in the generated modalities. This metric has become a standard in the evaluation of image generative models such as \textit{Generative Adversarial Networks} (GANs) \cite{goodfellow2014generative}. 
Inspired by the success of FID, we introduce the \textit{Fréchet Motion Distance} (FMD), a new objective metric to evaluate the quality and diversity of the synthesized human motions. There were similar previous attempts. \cite{xi2020generative} proposed applying FID concept to motion data, but there were no validating experiments such as how the proposed metric works for different motion noises. \textit{Fréchet Gesture Distance} (FGD) \cite{yoon2020speech} was proposed to evaluate gesture human motion, but it was bound to upper-body gesture. The code is available at \textcolor{blue}{\href{https://github.com/antmaio/FrechetMotionDistance}{https://github.com/antmaio/FrechetMotionDistance}}.
\begin{figure*}
    \centering
    \includegraphics[width=\textwidth, height=2.9cm]{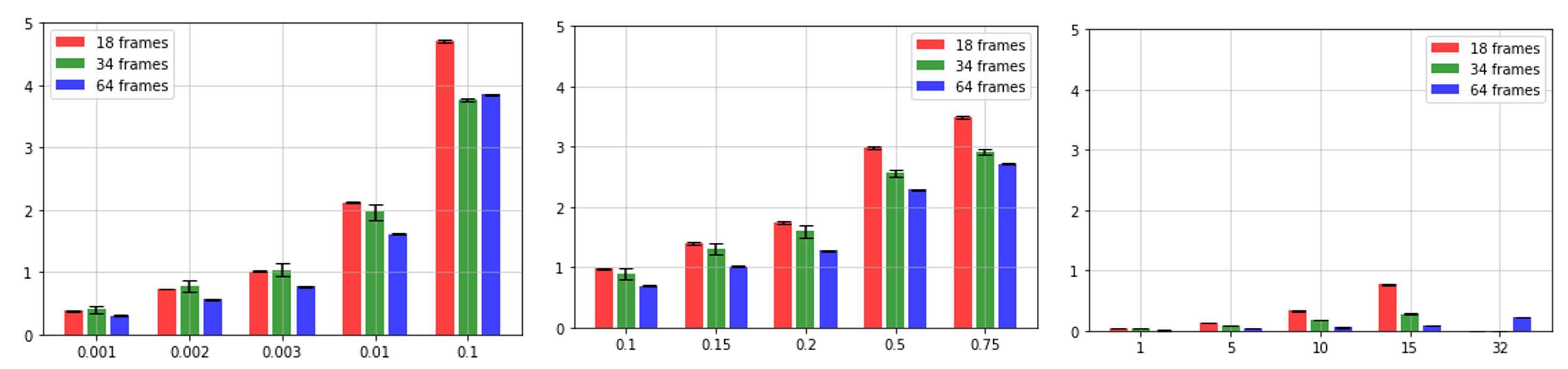}
    \caption{Mean FMD evolution with $\zeta$. We repeated the experiments 100 times for the same setting, and the error interval shows the standard deviation. (Left) Gaussian noise, (Mid) salt and pepper noise, (Right) temporal noise ($\zeta=32$ is only evaluated for a motion length of 64 frames). Low score means better synthesized motion. The FMD score is less sensitive to temporal noise but is consistent across the motion length.}
    \label{fig:res}
\end{figure*}
\section{Fréchet Motion Distance}
\label{sec:method}
Fig \ref{fig:overview} overviews the proposed method. An autoencoder encodes a motion dataset into a low-dimensional latent representation. Then, the Fréchet distance is computed between the latent spaces of a ground truth and a synthesized dataset by Equation \ref{eq:fid} which is the standard FID equation \cite{heusel2017gans}
\begin{equation}
    \label{eq:fid}
    FMD = || \mu_g - \mu_r ||^2_2 + Tr(\Sigma_r + \Sigma_g - 2\sqrt{\Sigma_g \Sigma_r}) 
\end{equation}
where ($\mu_g$, $\Sigma_g$) and ($\mu_r$, $\Sigma_r$) are respectively the mean and the covariance matrix of the synthesized and the ground truth motion distribution. This distance evaluates how close the synthesized motion distribution is to the ground truth dataset and assesses the quality and diversity of the generated motion. \cite{yoon2020speech} relies on a 1D convolutional autoencoder architecture to compute the metric since the temporal evolution of each cartesian position during the motion is considered as a 1D signal. These signals are converted into directional vectors of unit length between neighboring joints. However, this method does not evaluates efficiently the motion quality when varying the motion length. In constrast with \cite{yoon2020speech}, our method first transforms motion into an image representation. This transformation is based on the similarity between the dimension of the cartesian space and the RGB channels: $x$,$y$ and $z$ dimension play respectively the role of the red, blue and green channel on the image. The image width and height represents respectively the motion length and the number of directional vectors derived from the number of joints in the skeleton. The encoder part is a \textit{Resnet34}-based model \cite{he2015deep} and is pretrained with \textit{ImageNet} dataset \cite{deng2009imagenet}.

\section{Validation Experiments}
We used Human3.6M dataset \cite{h36m_pami} for our experiments. It is a large-scale motion capture dataset where 11 actors performed various actions (eating, taking photo, greeting, walking, etc) and this dataset is most widely used one in the motion synthesis research. Respectively 6 (actors 1,5-9) and 1 (actor 11) subjects constitute the training and testing set for the experiments. The proposed autoencoder is trained on the training set and learns the latent space by reconstructing the input image representing the motion. To validate our proposed metric, we need to find that it correctly measure the intensity of motion degradation. The test set is manually altered by adding noise samples with a fixed intensity factor $\zeta$ so that it can play the role of a synthetic motion dataset with artifacts. Then, FMD score is computed between the clean test set and the altered one. Three types of noise samples are considered in this work: Gaussian (spread across all the joints), salt-and-pepper (impulsive noise on some joints) and temporal noise. The temporal noise only add Gaussian noise samples on $\zeta$ frames to create temporal discontinuity inside a motion. 
\noindent
Moreover, FMD score should be consistent across the motion temporal length that is encoded. Indeed, if one wants to evaluate the quality of a synthesized motion with an arbitrary number of frames, the expected behavior of FMD is that the motion degradation must be captured regardless the motion length. In the experiments, the impact of the motion length on FMD score is studied by considering 18,34 and 64 frames as the length of the motion encoded into the latent space.   

\section{Results}
Figure \ref{fig:res} shows the evolution of FMD with $\zeta$. Since the FMD measures a distance between two distributions, a low FMD score means that the distributions of generated and ground truth motion are close. First, the score increases with the noise intensity, regardless of the motion length and the type of noise. This means that the latent space is sensitive to the intensity of the motion alteration. However, the evaluation gives significantly lower score considering the temporal noise degradation. That type of artifact has a less significant influence on the latent space distribution. The temporal discontinuities are not embodied efficiently into the latent spaces. 

\section{Conclusion}
This work presents a method to assess the performance of a motion generative model objectively and without requiring human subjects. The proposed FMD measures a Fréchet distance between the distributions of the synthesized and ground truth motion. We validated the metric with synthetic noise data, and the metric measured efficiently the motion degradation with Gaussian and salt and pepper noise. There is however room for improvements considering the temporal noise alteration, and we need further validation on synthesized motions from recent generative models. 


\bibliographystyle{ACM-Reference-Format}
\bibliography{sample-base}

\end{document}